\documentclass[12pt]{iopart}

\usepackage[T1]{fontenc} 
\usepackage{float}
\usepackage{graphicx}
\usepackage{subcaption}
\usepackage{caption}
\usepackage{url}
\begin{document}

\title{LLM-guided Chemical Process Optimization with a Multi-Agent Approach}

\author{Tong Zeng$^1$, Srivathsan Badrinarayanan$^1$, Janghoon Ock$^{1,5}$, Cheng-Kai Lai$^1$ and Amir Barati Farimani$^{2,1,3,4}$}

\address{$^1$ Department of Chemical Engineering, Carnegie Mellon University, Pittsburgh, PA 15213, USA}
\address{$^2$ Department of Mechanical Engineering, Carnegie Mellon University, Pittsburgh, PA 15213, USA}
\address{$^3$ Department of Biomedical Engineering, Carnegie Mellon University, Pittsburgh, PA 15213, USA}
\address{$^4$ Machine Learning Department, Carnegie Mellon University, Pittsburgh, PA 15213, USA}
\address{$^5$ Department of Chemical and Biomolecular Engineering, University of Nebraska--Lincoln, Lincoln, NE 68588, USA}
\ead{barati@cmu.edu}
\vspace{10pt}

\begin{abstract} 
Chemical process optimization is crucial to maximize production efficiency and economic performance. Optimization algorithms, including gradient-based solvers, numerical methods, and parameter grid searches, become impractical when operating constraints are ill-defined or unavailable, requiring engineers to rely on subjective heuristics to estimate feasible parameter ranges. To address this constraint definition bottleneck, we present a multi-agent framework of large language model (LLM) agents that autonomously infer operating constraints from minimal process descriptions, then collaboratively guide optimization using the inferred constraints. Our AutoGen-based agentic framework employs OpenAI's o3 model, with specialized agents for constraint generation, parameter validation, simulation execution, and optimization guidance. Through two phases: (i) autonomous constraint generation using embedded domain knowledge, and (ii) iterative multi-agent optimization, the framework eliminates the need for predefined operational bounds. Validated on the hydrodealkylation process across cost, yield, and yield-to-cost ratio metrics, the framework demonstrated competitive performance with conventional optimization methods while achieving a 31-fold reduction in wall-time relative to grid search, converging in under 20 minutes and requiring far fewer iterations to converge. Beyond computational efficiency, the framework's reasoning-guided search demonstrates sophisticated process understanding, correctly identifying utility trade-offs, and applying domain-informed heuristics. Unlike conventional optimization methods like Bayesian Optimization that require predefined constraints, our approach uniquely combines autonomous constraint generation with interpretable, reasoning-guided parameter exploration. Reproducibility analysis across five independent trials demonstrates consistent convergence behavior, while model comparison reveals that reasoning-capable LLM architectures (o3, o1) are essential for successful optimization, with standard models failing to converge effectively. This approach shows significant potential for optimization scenarios where operational constraints are poorly characterized or unavailable, particularly for emerging processes and retrofit applications.
\end{abstract}

\vspace{2pc}

\section{Introduction}

Efficient optimization of chemical processes is a central objective in process systems engineering, with direct implications for production efficiency and economic performance\cite{biegler2010nonlinear}. Various optimization approaches have been developed to address different aspects of process optimization challenges. Bayesian optimization (BO)\cite{snoek2012practicalbayesianoptimizationmachine, WANG2022100728} is one approach to black-box optimization, though it is computationally expensive and requires substantial function evaluations to converge. Gradient-based solvers remain the dominant methods in process optimization, including Interior Point OPTimizer (IPOPT)\cite{wachter2006implementation} and Sparse Nonlinear OPTimizer (SNOPT)\cite{gill2005snopt}, which depend on the smoothness and differentiability of the underlying objective function\cite{biegler2004retrospective}. When these requirements cannot be met, alternative methods such as evolutionary algorithms\cite{GROSS1998S229}, and derivative-free methods\cite{rios2013derivative} provide viable solutions. Furthermore, exhaustive search techniques\cite{edgar2001optimization} such as grid search systematically evaluate all combinations of discretized parameter values throughout the design space.  

However, all of these conventional optimization methods, including BO, require a complete and well-defined design space with described operating constraints. This requirement makes it difficult to apply these methods when feasible parameter ranges or process operating conditions are incomplete or unavailable. In these cases, engineers typically rely on heuristics or domain knowledge to estimate feasible ranges\cite{douglas1988, Himmelblau1996}. These heuristic approaches are highly subjective and may lead to variable optimization results. Mathematical simulation platforms such as IDAES\cite{lee2021idaes} and modeling libraries such as Pyomo\cite{bynum2021pyomo} offer flexible environments to build detailed process models and utilize IPOPT as its solver for scalability, speed. These platforms, however, still require users to define appropriate variable bounds and formulate smooth, well-posed objective functions before optimization. This constraint definition bottleneck significantly limits the application of optimization techniques across the board, particularly for non-expert users or for novel processes where operational configurations are not well-established.

Recent developments in large language models (LLMs) offer potential approaches to address these limitations. LLMs have demonstrated strong reasoning capabilities in complex problem-solving tasks\cite{wei2022chain, brown2020language, Xin2025} and effective adaptation to previously unseen domains through few-shot learning\cite{srivastava2023beyond}. Most remarkably, recent work has shown that LLMs can autonomously conduct scientific research, from hypothesis generation and experimental design to data analysis and manuscript writing\cite{lu2024ai}, demonstrating their potential for independent scientific reasoning and discovery. 

These general capabilities have translated into significant advances within chemical engineering applications\cite{adsorb_agent, matsci_agent, llm3dprint}. In synthesis planning, LLMs evaluate synthetic routes based on strategic requirements\cite{bran2025chemical}, suggesting that domain expertise traditionally requiring years of training could become more accessible through natural language interfaces. For material discovery, LLM-based agents such as ChemCrow\cite{bran2023chemcrow} have successfully planned and executed syntheses using automated laboratory platforms. For control system design\cite{guo2024controlagent, vyas2024autonomous, schweidtmann2024generative, balhorn2024toward, hirtreiter2023toward}, LLMs enable more flexible process control approaches without the need for specialized programming skills.

However, most existing efforts focus on isolated decision steps or downstream control, with limited exploration of LLM integration into steady-state simulation or process optimization pipelines, and autonomous constraint generation for steady-state process optimization using collaborative multi-agent LLM systems remains largely unexplored. This work addresses this gap by introducing an autonomous, multi-agent framework that integrates language model reasoning with process simulation for operating condition optimization. We tackle the challenge of autonomous constraint generation when feasible operating ranges are unknown or poorly defined. 

Our approach leverages the AutoGen platform\cite{wu2024autogen} with OpenAI's o3 model\cite{OpenAI2025o3} to create a multi-agent system where language model agents collaborate to both generate realistic constraints and iteratively optimize process parameters. The framework is evaluated on a real-world chemical flowsheet - the hydrodealkylation (HDA) process,
using IDAES\cite{lee2021idaes} simulations and assessed across three metrics: cost, yield, and yield-to-cost ratio. We selected IPOPT and grid search as representative benchmarks: IPOPT as a state-of-the-art gradient-based solver widely used in chemical process modeling, and grid search as a gradient-free exhaustive method that provides a global optimization baseline across the full parameter space. The results show comparable performance, while requiring fewer iterations and less computational time. This approach could significantly reduce the expertise barrier for process optimization and enable more widespread application of advanced optimization techniques in industrial settings, particularly for emerging processes or retrofitting existing operations where operational constraints are poorly characterized.

\section{Methodology}
\subsection{Framework Design}

We propose a multi-agent interactive system using AutoGen's GroupChat architecture\cite{wu2024autogen} that addresses fundamental limitations of single-agent approaches for complex optimization tasks. Individual agents maintain focused prompts for specific functions, preventing context dilution that occurs when lengthy multi-purpose prompts overwhelm LLM attention mechanisms. Each agent's specialized role enables direct function calling rather than complex output parsing, reducing implementation complexity and error potential. The modular design allows independent debugging and modification of optimization components, critical for iterative algorithm development and deployment flexibility.

\begin{figure}[!ht]
  \centering  \includegraphics[width=0.8\textwidth]{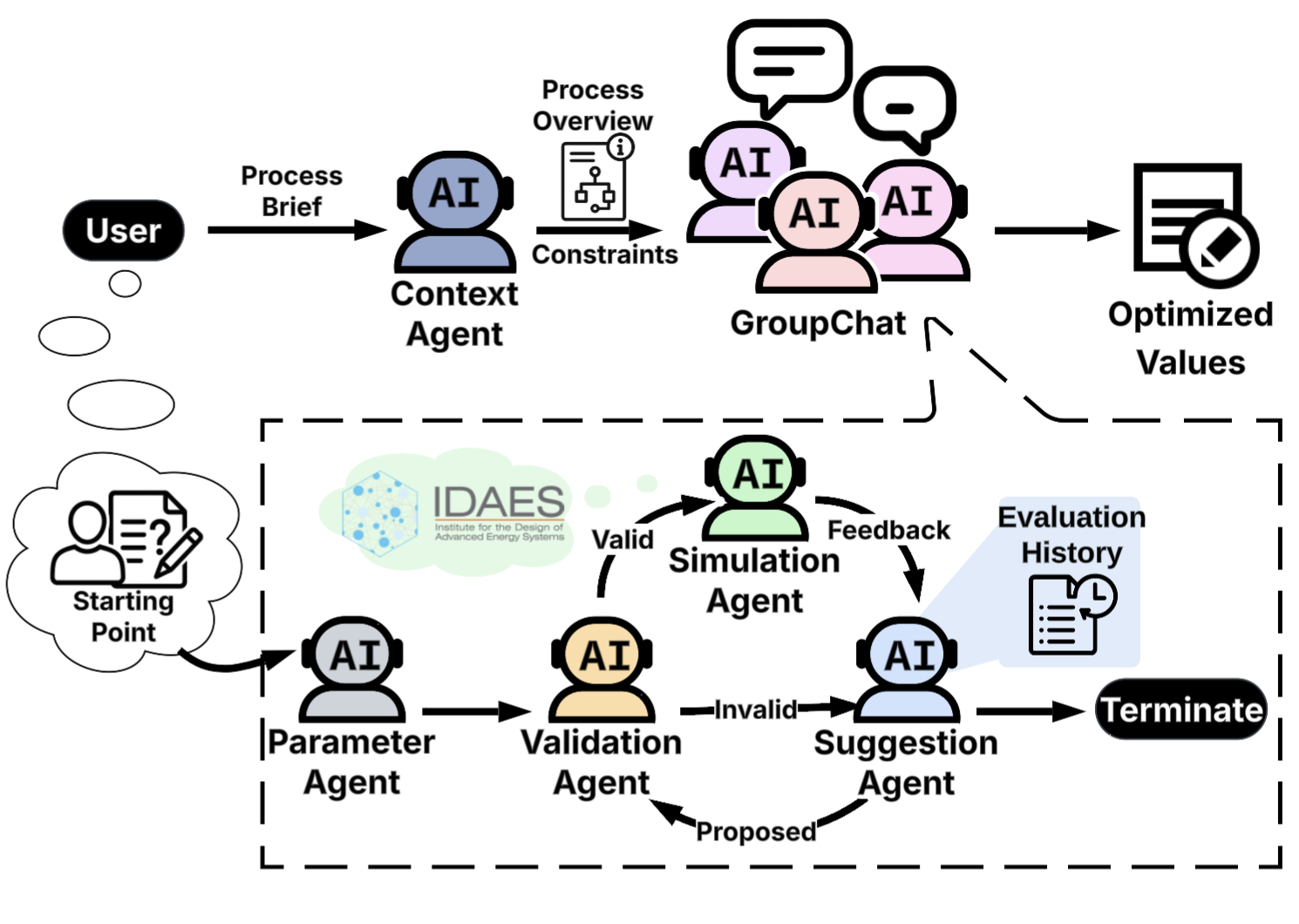}
  \caption{Multi-agent workflow of the LLM-based optimization framework. The ContextAgent generates constraints and detailed process overview from basic process info using domain knowledge, while specialized agents within the GroupChat environment handle parameter introduction, validation, optimization suggestions, and IDAES-based simulation evaluation through iterative refinement until optimal solutions are achieved.}
  \label{fig:flowsheet_autogen}
\end{figure}

Our framework consists of five specialized LLM agents, where each "agent" represents a specialized LLM instance with distinct prompts and function-calling capabilities designed for specific optimization tasks. The system operates through two phases: constraint generation followed by iterative optimization within a collaborative multi-agent environment, as shown in Figure~\ref{fig:flowsheet_autogen}. Throughout this process, autonomy refers to each agent's ability to reason, act, and iterate without human intervention, using only embedded domain knowledge, generated constraints, and shared optimization history.

The first phase involves the \textbf{ContextAgent} operating independently to infer realistic variable bounds and generate process context from given descriptions using embedded domain knowledge. This agent infers constraint ranges as lower and upper bounds for each decision variable and creates a detailed process overview text that provides contextual understanding for subsequent optimization agents (detailed process overview generated by ContextAgent shown in Supporting Information). These generated constraints and process overview serve as the foundation for the subsequent optimization phase.

The second phase uses the GroupChat environment to support coordinated interactions and shared memory between four specialized agents during optimization. Through iterative cycles of parameter proposal, validation, simulation, and refinement, the agents collaboratively navigate the constraint-defined parameter space toward optimal solutions.

Here, we document that all computational experiments in this study were performed on a MacBook Pro equipped with an Apple M1 chip (8-core, 16 GB RAM). The framework requires no model training or GPU resources, relying on API-based access to pre-trained language models.

\subsection{Agent Roles and Functions}

Within the GroupChat architecture, each agent maintains specialized responsibilities that coordinate to enable autonomous optimization. The optimization process operates through coordinated agent interactions in a specific sequence:

\textbf{ParameterAgent} initiates the process by receiving parameter-value pairs from users as initial starting points. These initial guesses can be arbitrary and may violate feasible bounds, as users typically lack knowledge of the constraint ranges previously generated by the ContextAgent.

\textbf{ValidationAgent} serves as the critical first checkpoint, evaluating each parameter set against the generated constraints and providing feedback about whether proposed values are acceptable or identifying which specific parameters need adjustment due to constraint violations. When invalid input is detected, the system automatically redirects to the SuggestionAgent for revision, creating a validation-correction loop that ensures all parameter sets remain within feasible bounds.

\textbf{SimulationAgent} executes process evaluation once parameters pass validation. This agent runs a custom objective function that interfaces with a pre-defined python-based IDAES\cite{lee2021idaes} simulation model (detailed IDAES implementation, thermodynamic properties, and kinetic parameters are provided in Supporting Information), calculating performance metrics such as cost, yield, and yield-to-cost ratios for each parameter configuration. Performance results are reported back to the SuggestionAgent, including any simulation convergence issues.

\begin{figure}[!ht]
  \centering  \includegraphics[width=0.8\textwidth]{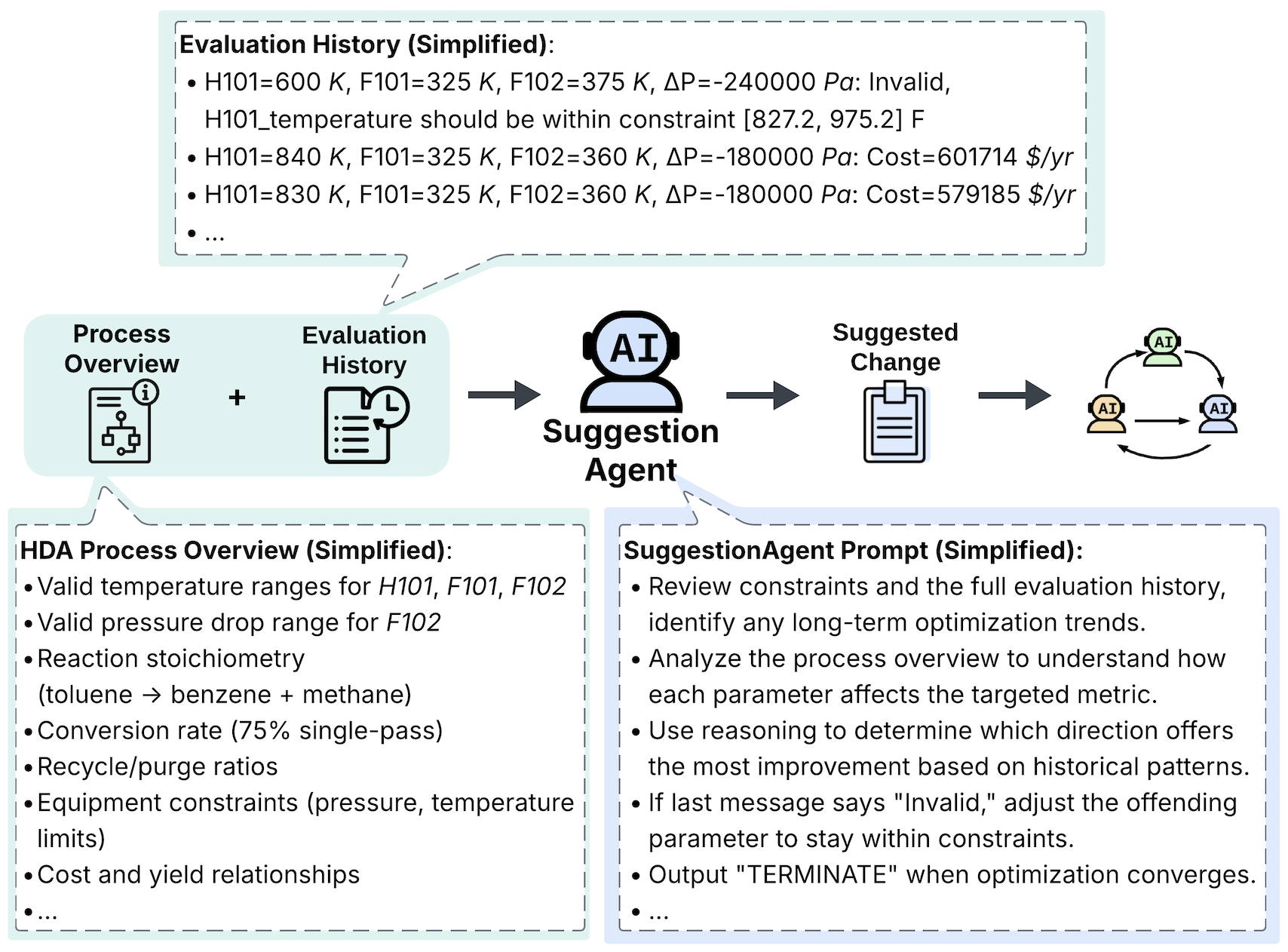}
  \caption{Visual illustration of the workflow of the SuggestionAgent. The agent receives three key inputs: (1) a structured prompt with optimization instructions, (2) process overview including valid constraint ranges, and (3) evaluation history of previous attempts. The SuggestionAgent analyzes these inputs to generate feasible parameter suggestions that satisfy all constraints, which are then validated and refined through iterative multi-agent interactions.}
    \label{fig:flowsheet_suggestion}
\end{figure}

\textbf{SuggestionAgent} serves as the optimization engine, maintaining a comprehensive record of the optimization history.  After each iteration, current parameter values and corresponding simulation results are appended to the agent's memory, providing direct access to both valid bounds and historical trial data. This accumulated knowledge enables the agent to refine and propose future parameter sets based on observed trends and constraint violations, completing the iterative optimization cycle. Figure~\ref{fig:flowsheet_suggestion} illustrates the SuggestionAgent workflow with simplified representations of the prompt instructions, process overview, and evaluation history, with the complete prompt provided in Supporting Information.

Agent behavior is controlled through carefully designed prompts that define roles, objectives, and output formats. The ContextAgent prompt incorporates engineering domain knowledge and formatting requirements to generate realistic operating constraints from process descriptions, while the SuggestionAgent prompt implements memory-guided optimization logic with constraint violation handling. Complete prompt templates are provided in Supporting Information, demonstrating the integration of process engineering heuristics with structured output requirements that enable autonomous constraint generation and iterative parameter refinement.

\subsection{Optimization Workflow}

The iterative optimization process operates through a precisely defined agent interaction sequence that ensures systematic parameter refinement. Agent interactions are coordinated through a customized selector function (detailed algorithm is provided in Supporting Information) that determines speaking order based on current speaker and message content, preventing redundant communications and ensuring logical workflow progression. 

The core optimization cycle follows this sequence: ParameterAgent introduces values $\rightarrow$ ValidationAgent checks feasibility $\rightarrow$ SimulationAgent evaluates performance $\rightarrow$ SuggestionAgent analyzes results and proposes improvements $\rightarrow$ ValidationAgent verifies new proposals. This cycle repeats autonomously until the SuggestionAgent detects convergence based on diminishing performance improvements across successive iterations.

This structured workflow design addresses three critical challenges in autonomous optimization: (1) constraint enforcement through mandatory validation checkpoints, (2) systematic exploration through memory-guided suggestions, and (3) autonomous termination through trend analysis. Unlike gradient-based methods that require mathematical derivatives or grid search that evaluates fixed discretizations, this agent-based approach leverages reasoning to intelligently navigate parameter spaces, adapting search strategy based on constraint violations and performance feedback.  

The framework's autonomous nature is demonstrated through self-determined termination when the SuggestionAgent detects diminishing returns based on simulation history. This approach reflects the agent's ability to prevent unnecessary computation while adapting to different problem characteristics, ensuring computational efficiency without sacrificing solution quality.

\subsection{Evaluation Metrics}

The SimulationAgent evaluates process performance using three metrics that capture the primary objectives of chemical process optimization: operating cost, product yield, and yield-to-cost ratio. 

Annual operating cost\cite{idaes_hda_flowsheet_2024} serves as the primary economic indicator, minimizing energy consumption through optimal utility management including heating costs and cooling costs for operating units:
\begin{equation}
\rm{Operating\ Cost} = \rm{Heating\ Cost} + \rm{Cooling\ Cost} \quad [\$/\rm{yr}]
\end{equation}
where the detailed heating and cooling cost calculations with utility coefficients are provided in Supporting Information.

Annual product yield represents production efficiency, maximizing production capacity by optimizing the absolute production rate: 
\begin{equation}
\rm{Yield} = \rm{F_{benzene,reactor}} \times 3600 \times 24 \times 365 \quad [mol/yr]
\end{equation}
where $F_{benzene,reactor}$ is explicitly defined as the molar flow rate of benzene at the reactor outlet (R101) in units of $mol/s$.

The yield-to-cost ratio provides a composite metric that balances these competing objectives, enabling evaluation of trade-offs between maximizing output and minimizing expenses:
\begin{equation}
\mbox{Yield--to--Cost Ratio} = \frac{\rm{Annual\ Yield}}{\rm{Operating\ Cost}} \quad [\rm{mol/\$}]
\end{equation}

These metrics were selected to cover industrial optimization priorities: economic performance (cost), production effectiveness (yield), and overall process efficiency (yield-to-cost ratio), allowing direct comparison with conventional optimization approaches across multiple performance dimensions. It is important to note that "Operating Cost" is defined as the sum of variable utility costs (Heating Cost + Cooling Cost) and thus excludes fixed charges or capital costs.

\section{Results and Discussion}
\subsection{Case Study Validation}
\begin{figure}[H]
  \centering  \includegraphics[width=0.8\textwidth]{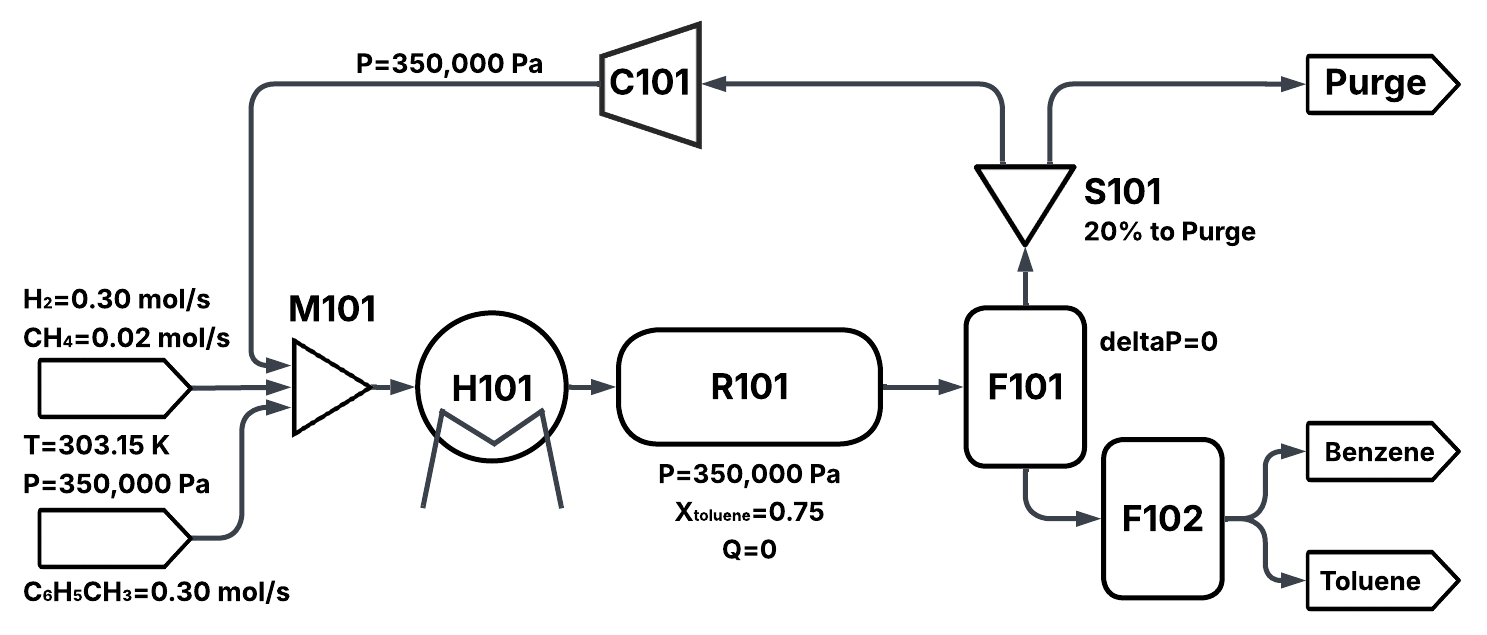}
  \caption{Flowsheet of the hydrodealkylation (HDA) process used for IDAES simulation and optimization study. The process consists of mixer M101, heater H101, reactor R101, flash separators F101 and F102, splitter S101, and compressor C101. Feed streams include hydrogen (0.30 mol/s), methane (0.02 mol/s) and toluene (0.30 mol/s) at 303.15 K and 350,000 Pa. The reactor operates at 350,000 Pa with 75\% toluene conversion, followed by separation units that produce benzene and toluene product streams, with 20\% of the overhead stream purged through splitter S101.}
  \label{fig:flowsheet_hda}
\end{figure}

To validate our multi-agent framework's constraint generation and optimization capabilities, we demonstrate its performance using the HDA process, which shows the constraint-limited optimization challenge our approach addresses. This process converts toluene to benzene through connected unit operations, but as shown in Figure~\ref{fig:flowsheet_hda}, only basic information is available: feed conditions, conversion rates, and equipment configuration. No operating temperature ranges, pressure limits, or feasible parameter bounds are specified. This process represents a realistic industrial scenario where detailed operational constraints are often unavailable, making it an ideal test case for autonomous constraint generation. The optimization challenge becomes apparent when attempting to apply traditional methods to this process.

Mathematical optimization methods like IPOPT or grid search cannot be directly applied without explicit constraint sets. This limitation demonstrates the practical need for our ContextAgent, which infers variable bounds using embedded process knowledge. Given this constraint generation capability, we formulate the optimization problem as follows:
\begin{equation}
\rm{optimize\ } \phi(x) \quad \rm{subject\ to} \quad x_i^L \leq x_i \leq x_i^U, \quad \forall i = 1,\dots,4
\end{equation}
where $\phi(x)$ denotes the objective function (e.g. yield, cost, or yield-to-cost ratio) evaluated by IDAES simulations, while $x^L$ and $x^U$ are variable bounds derived from LLM-generated domain knowledge. The vector $x = (x_1, x_2, x_3, x_4)$ represents the four key decision variables: H101 outlet temperature, F101 outlet temperature, F102 outlet temperature, and F102 pressure drop. The implementation of the IDAES model uses ideal vapor-liquid equilibrium relationships for thermodynamic calculations. The reactor operates with fixed 75\% toluene conversion using stoichiometric reaction constraints (complete process description, feed specifications, and definitions of objective functions are provided in Supporting Information).

While BO achieves sample efficiency when optimizing well-behaved continuous functions with well-defined constraints\cite{snoek2012practicalbayesianoptimizationmachine} \cite{WANG2022100728}, our problem presents characteristics that align poorly with BO's assumptions. First, constraint generation is part of the optimization problem. Unlike hyperparameter tuning where valid ranges are known, chemical process optimization requires determining feasible operating boundaries based on thermodynamic and kinetic principles as well as equipment limitations. Second, engineers need to understand why specific operating conditions are recommended for safety validation and regulatory compliance. LLMs provide natural language explanations grounded in chemical engineering principles, allowing them to justify parameter suggestions (e.g., 'Increase temperature to 350 \textdegree{}C to improve reaction kinetics while staying below thermal decomposition limits') and identify violations of physical laws or impractical operating regimes (e.g., conditions causing unintended phase changes or exceeding material temperature limits). In contrast, BO's parameter selection lacks interpretable explanations and cannot leverage encoded domain knowledge to validate recommendations.

\subsection{Constraint Generation Quality}
To assess the reliability and consistency of autonomous constraint generation, the ContextAgent's constraint generation capability was evaluated across five independent trials using identical prompts to ensure representative constraint generation and minimize stochastic variance. The coefficient of variation (CV) was calculated to quantify the consistency of constraint generation across trials. For each variable $i$, the CV was computed as the average of the lower-bound and upper-bound CVs:

$$
CV_i = \frac{1}{2}\left(\frac{\sigma_{L,i}}{\mu_{L,i}} + \frac{\sigma_{U,i}}{\mu_{U,i}}\right) \times 100\%
$$

where $\sigma_{L,i}$ and $\mu_{L,i}$ are the standard deviation and mean of the lower bounds, and $\sigma_{U,i}$ and $\mu_{U,i}$ are the standard deviation and mean of the upper bounds across the five trials. Table~\ref{tbl:constraint_avg} presents the individual trial results, average constraint ranges, and CV values for the operating conditions across key unit operations H101, F101, and F102. The average ranges were used for subsequent optimization performance comparisons.

\begin{table}[H]
\centering
\caption{ContextAgent Constraint Generation Results and Average Constraint Range (with Variation) Across Five Independent Trials}
\label{tbl:constraint_avg}
\scriptsize
\renewcommand{\arraystretch}{1.1}
\resizebox{\textwidth}{!}{%
\begin{tabular}{l|ccccc|c}
  \hline
  \textbf{Unit Variable} & \textbf{Trial 1} & \textbf{Trial 2} & \textbf{Trial 3} & \textbf{Trial 4} & \textbf{Trial 5} & \textbf{Average Range (CV (\%) $\downarrow$
)}\\
  \hline
  H101 Temp (K)          & [870, 950]       & [873, 973]       & [820, 980]       & [800, 950]       & [773, 1023]      & [827.2, 975.2] (4.18) \\
  F101 Temp (K)          & [310, 330]       & [300, 340]       & [290, 390]       & [330, 420]       & [298, 368]       & [305.6, 369.6] (7.49) \\
  F102 Temp (K)          & [300, 340]       & [330, 380]       & [280, 340]       & [300, 360]       & [323, 413]       & [306.6, 366.6] (7.48) \\
  F102 $\Delta P$ (Pa)   & [-200000, -50000] & [-200000, -50000] & [-240000, -10000] & [-200000, -10000] & [-220000, -20000] & [-212000, -28000] (40.82) \\
  \hline
\end{tabular}
}
\end{table}

As expected from the stochastic nature of LLM responses, constraint generation showed variability across trials, as detailed in the Supporting Information. For the critical H101 reactor temperature, bounds ranged from [773, 1023] K to [870, 950] K across trials, with range widths varying from 80 K to 250 K. Flash separator temperatures showed similar variability, with F101 bounds spanning from [290, 390] K to [330, 420] K, and F102 bounds from [280, 340] K to [323, 413] K. Pressure drop constraints were more consistent, with all trials generating similar lower bounds (-240000 to -200000 Pa) but varying upper bounds (-10000 to -50000 Pa). While this variability might initially appear concerning, it actually demonstrates the framework's robust constraint generation capability. The key insight is not perfect consistency, but rather the consistent generation of industrially viable bounds that enable optimization to proceed when traditional methods cannot be applied due to missing constraint information.

Quantitatively, the coefficient of variation (CV) analysis showed low variability for temperature-related constraints, below $10\%$, but substantially higher variability for the pressure drop constraint, around $40\%$. This indicates that the temperature bounds were generated more consistently across trials, whereas the pressure constraint was more sensitive to prompt stochasticity and less tightly grounded in the provided textual context.

Importantly, despite this observable variation across trials, the ContextAgent consistently generated realistic and engineering-appropriate bounds. For the critical H101 reactor temperature, the average generated range of [827.2, 975.2] K shows a good alignment with industrial practices reported at [773, 977] K\cite{Eldood2008, douglas1988}. Although specific industrial ranges for the flash separator temperatures and pressure drop are not readily available in the literature, the close correspondence between agent-generated reactor constraints and established industrial practice demonstrates the agent's ability to extract field-consistent operational conditions using only basic process information and embedded domain knowledge.

\subsection{Performance Comparison}
\textbf{Optimization Results: }To assess the effectiveness of our approach, we compared the LLM framework's optimization performance against IPOPT (detailed algorithm description is provided in Supporting Information) and grid search benchmarks using the averaged ContextAgent-generated constraints. Grid search was implemented with 10 discretization points per variable, resulting in 10,000 total parameter combinations (10×10×10×10) evaluated across the four decision variables. The achievement metric was calculated to quantify each method’s optimization performance relative to the grid search baseline. For each method, the achievement was computed as the ratio of the objective value obtained by the evaluated method to that obtained by the grid search benchmark: 
$$ \mathrm{Achievement} = \frac{V_{\mathrm{method}}}{V_{\mathrm{grid}}} \times 100\%$$
where $V_{\mathrm{method}}$ and $V_{\mathrm{grid}}$ are the objective values achieved by the evaluated optimization method and the grid search baseline, respectively. 
Table~\ref{tbl:averaged} summarizes the optimization results across the three evaluation metrics as defined earlier.

\begin{table}[H]
\centering
\caption{Comparison of IPOPT, LLM and Grid Search Optimization Results.}
\label{tbl:averaged}
\renewcommand{\arraystretch}{1.0}
\small
\resizebox{\textwidth}{!}{%
\begin{tabular}{llcccccc}
    \hline
    \textbf{Metric} & \textbf{Method} & \textbf{\shortstack{H101\\Temp (K)}} & \textbf{\shortstack{F101\\Temp (K)}} & \textbf{\shortstack{F102\\Temp (K)}} & \textbf{\shortstack{F102\\$\Delta P$ (Pa)}} & \textbf{Value} & \textbf{\shortstack{Achievement\\(\%) ($\uparrow$)}} \\
    \hline
    Cost (\$/yr)$^{a}$ & LLM  & 830.0 & 325.0 & 360.0 & -180000 & $5.792 \times 10^5$ & 97.72 \\
    & IPOPT       & 827.2 & 305.6 & 306.6 &  -28314 & \textbf{5.660$\times$10$^5$} & 100.00 \\
    & Grid Search & 827.2 & 305.6 & 306.6 &  -48444 & \textbf{5.660$\times$10$^5$} & 100.00 \\
    \hline
    Yield (mol/yr)$^b$ & LLM         & 975.0 & 369.5 & 366.0 & -180000 & $1.644 \times 10^7$ & 93.94 \\
      & IPOPT       & 857.5 & 369.6 & 344.4 & -130340 & $1.565 \times 10^7$ & 89.43 \\
      & Grid Search & 958.8 & 369.6 & 339.9 &  -89333 & \textbf{1.750$\times$10$^7$} & 100.00 \\
    \hline
    Yield/Cost ($\mathrm{mol/\$})^{b}$ & LLM & 827.2 & 369.0 & 306.7 &  -28000 & 17.23 & 97.79 \\
      & IPOPT       & 827.2 & 369.6 & 306.6 & -121039 &  \textbf{17.62}& 100.00 \\
      & Grid Search & 827.2 & 369.6 & 306.6 &  -68889 & \textbf{17.62} & 100.00 \\
    \hline
\end{tabular}
}
\vspace{1mm}
\footnotesize$^a$ Lower is better. $^b$ Higher is better.
\end{table}

\textbf{Performance Analysis: }The results shown above demonstrate competitive performance across all three metrics. Grid search consistently achieves the best solutions, benefiting from exhaustive evaluation of the discretized parameter space. However, the LLM-based optimization framework demonstrates comparable or better performance while being dramatically faster than conventional methods. Table~\ref{tbl:convergence} shows the number of iterations required for each optimization approach across different metrics.

\begin{table}[H]
\centering
\caption{Optimization Performance Comparison}
\label{tbl:performance}
\footnotesize
\begin{minipage}{0.48\textwidth}
\centering
\subcaption{Convergence Speed (Iterations)}
\label{tbl:convergence}
\begin{tabular}{lcc}
  \hline
  \textbf{Metric} & \textbf{LLM} & \textbf{IPOPT} \\
  \hline
  Cost (\$/yr)        & 21 & 35   \\
  Yield (mol/yr)       & 26 & 81   \\
  Yield/Cost (\rm{mol/\$})   & 43 & 46   \\
  \hline
\end{tabular}
\end{minipage}
\hfill
\begin{minipage}{0.48\textwidth}
\centering
\subcaption{Optimization Wall Time (Hours)}
\label{tbl:runtime}
\begin{tabular}{lcc}
  \hline
  \textbf{Metric} & \textbf{LLM} & \textbf{Grid Search} \\
  \hline
  Cost (\$/yr)        & 0.17  & 10.50 \\
  Yield (mol/yr)       & 0.20  & 10.50 \\
  Yield/Cost (\rm{mol/\$})  & 0.33  & 10.50 \\
  \hline
\end{tabular}
\end{minipage}
\end{table}

The LLM framework consistently required fewer iterations than IPOPT. Most notably, for yield optimization, IPOPT required 81 iterations compared to only 26 for the LLM approach, which is a 3-fold reduction. The time efficiency gains over exhaustive methods are even more dramatic. Table~\ref{tbl:runtime} shows the LLM framework completing optimization in under 20 minutes across all metrics, compared to grid search's fixed 10.5-hour requirement, which is a 31-fold speedup. While grid search must evaluate all 10,000 parameter combinations by design, the LLM framework converges in around 20-45 iterations through intelligent search strategy adaptation. 

Despite requiring significantly fewer iterations and computational time, the LLM framework achieves competitive solution quality. IPOPT performs marginally better in cost minimization and yield-to-cost ratio optimization, while the LLM framework outperforms IPOPT in yield maximization. These modest performance differences demonstrate the framework's feasibility for constraint-limited optimization problems, particularly given its substantial computational advantages. The framework's ability to balance solution quality with computational efficiency makes it especially valuable for scenarios requiring rapid optimization or when computational resources are limited.

To assess convergence behavior and reproducibility, we conducted five independent cost minimization runs using identical constraints and process descriptions. The framework demonstrated rapid convergence across runs, indicating consistent optimization behavior. Detailed convergence analysis are provided in Supporting Information.

\textbf{Reasoning-Guided Parameter Exploration: }To examine the agent's internal decision-making logic, we conducted a separate, additional trial with relaxed verbosity constraints. For cost minimization, the agent demonstrated sophisticated process understanding as shown below:
\begin{quote}
\textit{"Keep the reactor charge just inside the allowable window – the nearer-to-minimum H101 temperature ($\approx 830 K$) cuts furnace duty and fuel. Push both flash drums to the warm end of their ranges; warmer condensers mean less refrigeration/brine duty. Relax the second-flash pressure drop toward the mechanical minimum (--30 kPa). Throttling is essentially "free," so choosing the smallest $\Delta P$ avoids excessive vapour generation that would otherwise have to be recompressed or condensed downstream."}
\end{quote}

This reasoning demonstrates three key capabilities: correct identification of heating-cooling utility trade-offs, qualitative understanding of operating condition impacts on energy consumption, and application of domain-informed heuristics for parameter selection. The agent's ability to articulate process engineering principles while making parameter adjustments distinguishes this approach from conventional numerical optimizers. While this local cost-minimization strategy may not directly lead to the global optimum, it provides efficient search directions that can be refined through subsequent iterations.

Table~\ref{tbl:optimization_cost} demonstrates how the agent's reasoning translates into systematic parameter adjustments during cost minimization. The optimization process begins with arbitrary user-provided initial guesses that may violate the autonomously generated constraints. As shown, the initial H101 temperature of 600 K violates the average generated range [827.2, 975.2] K, resulting in "Invalid" status until the ValidationAgent-SuggestionAgent loop produces the first feasible solution at iteration 1. The progression reveals the agent's ability to systematically test different operating strategies and adapt based on performance feedback. Early parameter exploration shows varying cost performance as the agent tests different operating strategies, while subsequent iterations show the framework's convergence toward more cost-effective operating conditions. By iteration 10, the agent identifies an optimal reactor temperature of 830 K that substantially reduces operating costs, with later iterations focusing on fine-tuning secondary variables while maintaining this temperature setting. The framework's efficiency stems from this reasoning-guided approach, enabling convergence in 20-45 iterations compared to thousands required by exhaustive methods, while maintaining solution quality through systematic constraint enforcement and trend analysis.

\begin{table}[H]
\centering
\caption{Systematic parameter exploration during operating cost minimization.}
\label{tbl:optimization_cost}
\begin{tabular}{llccccc}
\hline
\textbf{Iteration} & \textbf{\shortstack{H101\\Temp (K)}} & \textbf{\shortstack{F101\\Temp (K)}} & \textbf{\shortstack{F102\\Temp (K)}} & \textbf{\shortstack{F102 \\$\Delta P$ (Pa)}} & \textbf{Cost (\$/yr)} \\
\hline
User Initial Guess  & 600.0 & 325.0 & 375.0 & -240000 & Invalid \\
1  & 840.0 & 325.0 & 360.0 & -180000 & 601715 \\
2  & 830.0 & 330.0 & 363.0 & -160000 & 617356 \\
6  & 840.0 & 330.0 & 357.0 & -175000 & 632627 \\
10  & 830.0 & 325.0 & 360.0 & -180000 & 579186 \\
$\vdots$ & $\vdots$ & $\vdots$ & $\vdots$ & $\vdots$ & $\vdots$ \\
\hline
\end{tabular}
\end{table}

\textbf{Framework Robustness Across Models: }The reasoning capabilities demonstrated above are not universal across all LLMs. To assess the framework's robustness and reproducibility, we evaluated the cost minimization task using four different models: OpenAI's o3 and o1 (reasoning models), and GPT-4o and GPT-4.1 (standard models). All models received identical prompts, constraints, and access to optimization history, isolating model reasoning capability as the sole variable.

\begin{figure}[H]
  \centering
  \includegraphics[width=0.9\textwidth]{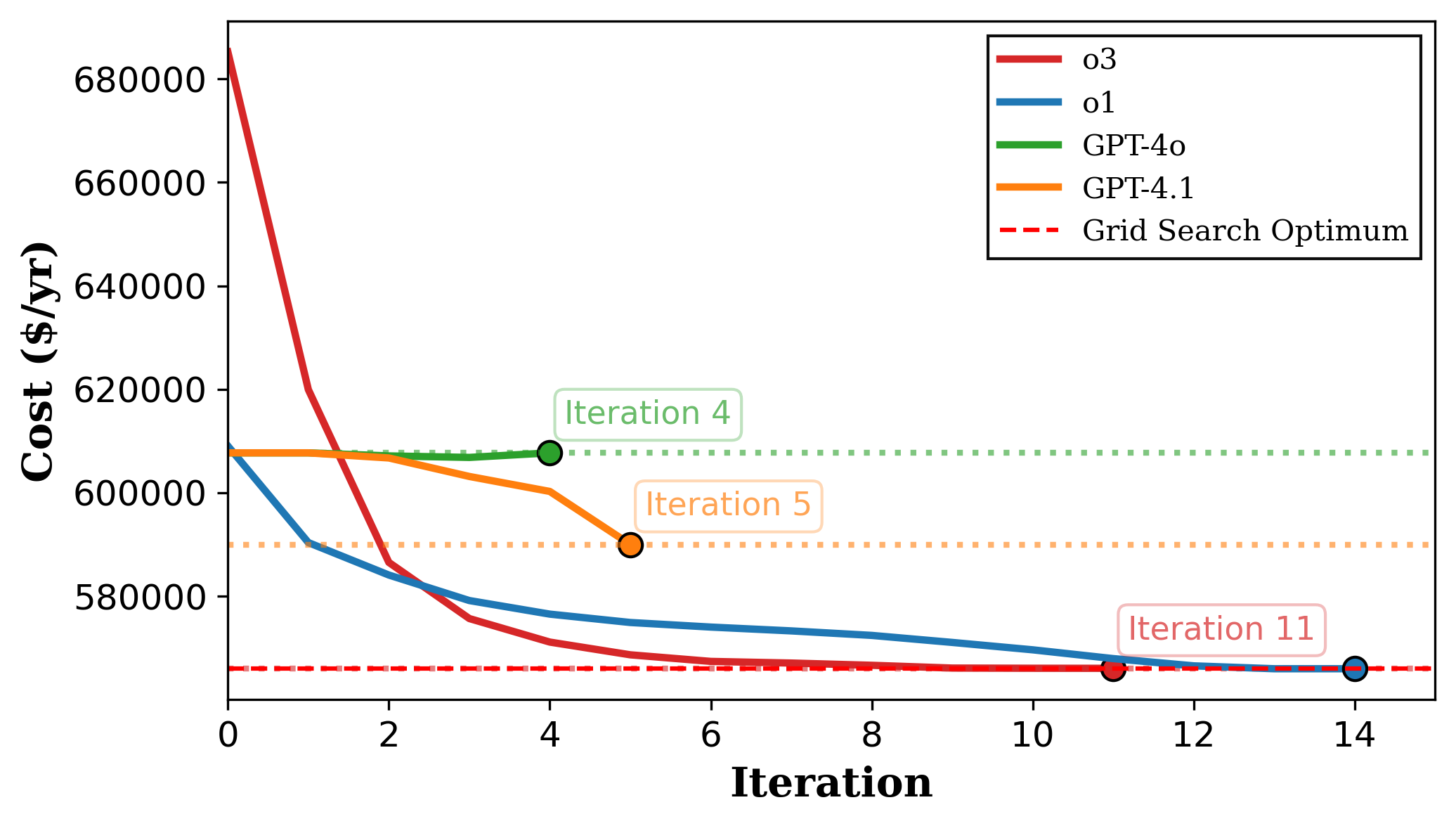}
  \caption{Convergence comparison across different LLM architectures for cost minimization. Solid dots indicate where models terminated, with dashed lines extending to show their final achieved cost for comparison. Red dashed line represents grid search's optimal cost solution of \$$5.660 \times 10^5/yr$. Reasoning models (o3 and o1) successfully converge, with o3 requiring 11 iterations and o1 requiring 14 iterations. Standard models (GPT-4o and GPT-4.1) terminate prematurely after 4 to 5 iterations without effective learning.}
  \label{fig:model_comparison}
\end{figure}

Figure~\ref{fig:model_comparison} reveals a significant performance gap between specialized reasoning models and general models. Only o3 and o1 successfully converged to solutions comparable with the grid-search benchmark, with o3 demonstrating superior efficiency by reaching the benchmark solution in 11 iterations compared to o1's 14 iterations. In contrast, GPT-4o and GPT-4.1 failed to converge, stalling at suboptimal solutions after merely 4 and 5 iterations respectively, despite having access to the same information and agent tools.

This performance gap highlights a critical requirement for autonomous optimization: successful operation demands more than access to domain knowledge or the ability to follow instructions. It requires sustained reasoning capabilities that enable multi-step planning, constraint satisfaction, and iterative refinement based on feedback. The superior performance of o3 over o1, despite both being reasoning models, suggests that advances in reasoning depth and consistency directly translate to optimization efficiency. This finding has important implications for framework deployment: while the approach demonstrates feasibility for autonomous constraint generation and optimization, its practical application currently requires access to advanced reasoning-capable models. Future work should investigate strategies to enhance robustness across model architectures, potentially through improved prompt engineering, hybrid approaches combining multiple models, or integration with conventional optimization methods for critical decision steps.

\textbf{Methodological Advantages: }The LLM approach does not require objective functions to be differentiable or continuous, potentially making it applicable to optimization problems with discontinuous or non-differentiable objectives. The scalability advantages become particularly evident when considering problem dimensionality. Grid search suffers from exponentially increasing computational demands as the number of decision variables grows (scaling as $n^d$ where n is discretization points and d is number of variables), with our four-variable implementation already requiring over 10 hours wall time despite using only 10 discretization points per variable. Our reasoning-guided approach intelligently navigates the parameter space without exhaustive enumeration, suggesting the framework has potential to handle higher-dimensional problems. 

Beyond scalability, the framework's process-agnostic design positions it well for application across diverse chemical processes. The ContextAgent requires only textual process descriptions to generate constraints, while the SimulationAgent interfaces with any IDAES-compatible model. Extending validation to reactive systems and energy systems would strengthen evidence of broad applicability, likely involving prompt customization to incorporate domain-specific heuristics relevant to each process class. This flexibility in prompt customization, supported by the pre-trained LLMs' embedded domain knowledge, enables extensions across diverse engineering domains and represents valuable opportunities for future work.

\section{Conclusions}

This work demonstrates that LLM agents can autonomously solve the constraint definition bottleneck that limits traditional optimization methods. By generating engineering-appropriate operating bounds from minimal process descriptions and iteratively optimizing within those bounds, our multi-agent framework enables process optimization in scenarios previously requiring extensive domain expertise or manual constraint specification.

The framework's primary innovation lies in autonomous constraint generation combined with collaborative multi-agent optimization. By leveraging embedded domain knowledge to infer engineering-appropriate operating bounds for unit operations, the system enables optimization in scenarios where traditional methods cannot be directly applied. The framework's flexibility in handling black-box and non-differentiable objective functions, combined with its ability to provide natural language explanations for parameter choices, addresses key limitations of conventional optimizers. Validation on the HDA process demonstrated competitive performance against IPOPT and grid search across multiple objectives, with significant computational efficiency gains achieving convergence in under 20 minutes compared to over 10 hours for grid search. Beyond computational metrics, the framework's reasoning capabilities provide interpretable optimization trajectories, with agents explicitly articulating process trade-offs and domain-informed parameter selection strategies that enhance engineer trust and facilitate safety validation.

This work establishes a foundation for autonomous process optimization that could significantly reduce expertise barriers and enable wider application of advanced optimization techniques. The demonstrated feasibility on the HDA process provides a proof-of-concept for the multi-agent approach, opening pathways for extending functionality toward autonomous process modeling and flowsheet synthesis. Key opportunities for advancement include integration with process databases for better constraint generation, and development of approaches that combine LLM reasoning with conventional optimization methods for improved robustness.

\section*{Code Availability}

All code used in this study is openly available at:  \\
\url{https://github.com/tongzeng24/ProcessAgent}

\section*{Supporting Information}

The Supporting Information includes multi-agent algorithm specifications, complete agent prompt templates, ContextAgent process understanding demonstrations, constraint generation approach comparisons, and detailed HDA process implementation including feed specifications, economic objective function formulations with utility cost coefficients, operational constraints, and IDAES modeling details.

\section*{Author Contributions}

T.Z., S.B., J.O., C.L and A.B.F. conceived the project. T.Z. developed the core multi-agent optimization framework, and led the manuscript preparation. S.B. contributed to the development of the constraint generation modules, refined the core optimization algorithm, and assisted with manuscript writing. J.O. provided key conceptual input on the design of the overall framework and optimization strategy. C.L. contributed to the implementation and validation of the IDAES-based simulations. All authors reviewed and approved the final manuscript. A.B.F. supervised the project.

\section*{References}

\bibliographystyle{unsrt}
\bibliography{references}

\end{document}


\title{Supporting Information: LLM-guided Chemical Process Optimization with a Multi-Agent Approach}

\author{Tong Zeng$^1$, Srivathsan Badrinarayanan$^1$, Janghoon Ock$^1$, Cheng-Kai Lai$^1$ and Amir Barati Farimani$^{2,1,3,4}$}

\address{$^1$ Department of Chemical Engineering, Carnegie Mellon University, PA 15213, USA}
\address{$^2$ Department of Mechanical Engineering, Carnegie Mellon University, PA 15213, USA}
\address{$^3$ Department of Biomedical Engineering, Carnegie Mellon University, PA 15213, USA}
\address{$^4$ Machine Learning Department, Carnegie Mellon University, PA 15213, USA}
\ead{barati@cmu.edu}

\section{Multi-Agent Framework Implementation}

\subsection{Agent Coordination Logic}
The framework uses a two-phase architecture where the ContextAgent independently generates operating constraints and process overviews, followed by iterative optimization through multi-agent collaboration. In the second phase, agent interactions are coordinated through a customized selector function that determines speaking order based on the source and message content of the most recent message.  The selector function implements a rule-based routing system that directs speaking flow between specialized agents. Constraint violations trigger redirection to the SuggestionAgent for parameter correction, while valid parameters proceed to the SimulationAgent for evaluation. The system achieves autonomous termination when the SuggestionAgent signals convergence. To minimize stochastic variance, the ContextAgent generates operating condition constraints five times with results averaged to establish the final constraints for optimization. The complete multi-agent optimization algorithm integrating both constraint generation and iterative optimization phases is presented below in Algorithm~\ref{alg:multiagent}. \\

\newcommand{\agent}[1]{\texttt{#1}}

\begin{algorithm}[H]
\caption{Multi-Agent Optimization Framework}
\label{alg:multiagent}
\textbf{Phase 1: Autonomous Constraint Generation}\\
Generate five independent constraint sets using \agent{ContextAgent}\\
Compute average bounds for each decision variable\\[0.5em]

\textbf{Phase 2: Collaborative Optimization}\\
\agent{ParameterAgent} introduces initial parameter values\\
\While{convergence not achieved}{
    \agent{ValidationAgent} evaluates parameter feasibility against constraints\\
    \eIf{parameters violate constraints}{
        \agent{SuggestionAgent} proposes corrected values\\
    }{
        \agent{SimulationAgent} evaluates objective function using IDAES\\
        Update optimization history in shared memory\\
        \agent{SuggestionAgent} analyzes trends and proposes next iteration\\
        \If{\agent{SuggestionAgent} detects convergence}{
            Terminate optimization\\
        }
    }
}
\textbf{Output:} Optimal parameter configuration
\end{algorithm}

\subsection{Agent Prompts}

Agent behavior is controlled through carefully designed prompts that define roles, objectives, and output formats. The following sections present the complete prompt templates of the key agents that guide autonomous constraint generation and optimization decisions.

\subsubsection{ContextAgent Prompt}
The optimization framework relies on prompt-engineered language model agents to reason about constraints and optimization strategies. Two central agents, ContextAgent and SuggestionAgent enable autonomous constraint generation and iterative parameter refinement. The following sections present the full prompt templates used to guide these agents' behavior.

The ContextAgent is responsible for generating realistic operating constraints based on a natural language process description. It interprets flowsheet structure, design basis, and engineering heuristics to infer valid bounds for decision variables. Its output is formatted as a structured JSON object containing both a markdown-style overview and numerical constraint ranges. The complete prompt for ContextAgent is:
\begin{lstlisting}[caption={Complete ContextAgent Prompt}, label={lst:context_prompt}]
You are a senior process-design engineer in HDA process to produce benzene from toluene.

OBJECTIVE
(1) Write a structured and technically complete process overview of the hydrodealkylation (HDA) flowsheet.
(2) Provide numerically realistic operational constraints for the variables listed below.

FORMATTING RULES
- Use SI units (K, Pa).
- If a limit is fixed in the Design Basis, copy it verbatim.
- No speculative extremes or lab-only conditions.
- For Flash F102 DeltaPressure, respect physical constraints:
    - The vessel is pressure-rated and the allowable maximum pressure drop must not exceed 240000 Pa in magnitude.
    - Output values must lie within [-250000, 0] Pa.
- The pressure of Flash F102 must not influence or constrain the allowable temperature range of Heater H101. These units operate independently in terms of pressure-temperature limitations.

DESIGN BASIS (KNOWN & FIXED)
- Feed T = 303.2 K; P = 350 kPa
- Mixer M101: blends Streams A, B, and recycle gas; no duty.
- Heater H101: adiabatic R101 feed
- Reactor R101: single-pass toluene conversion fixed at 75%; Q = 0
- Flash F101: approx. 350 kPa; DeltaP approx. 0
- Splitter S101: 20% purge / 80% recycle (fixed)
- Compressor C101: boosts recycle vapour back to 350 kPa (isothermal)
- Flash F102: low-pressure, benzene-rich overhead / toluene-rich bottoms

FLOW SHEET (for context only)
- Reaction (gas-phase, exothermic): C6H5CH3 + H2 -> C6H6 + CH4
- Feeds at 303.2 K & 350 kPa:
    - Stream A: 0.30 mol/s H2 + 0.02 mol/s CH4
    - Stream B: 0.30 mol/s toluene

VARIABLES REQUIRING CONSTRAINTS
- Heater H101 Temperature
- Flash F101 Temperature
- Flash F102 Temperature
- Flash F102 DeltaPressure (include sign)

Return valid JSON with exactly these keys:
{
  "process_overview": "<markdown overview goes here>",
  "constraints": [
    {"variable": "<Variable name>", "range": [<lower>, <upper>], "unit": "<unit>"},
    ...
  ]
}
\end{lstlisting}

\subsubsection{SuggestionAgent Prompt}

The SuggestionAgent performs the core optimization task. It analyzes previous simulation outcomes and constraint violations to propose parameter updates that improve a specified metric (e.g., cost, yield). The agent maintains memory of successful configurations and adapts its strategy based on trend direction and validator feedback. The complete prompt for SuggestionAgent is:
\begin{lstlisting}[caption={Complete SuggestionAgent Prompt}, label={lst:suggestion_prompt}]
You are SuggestionAgent.

What you can see
----------------
1. constraint_memory  (chronological)
- First lines: static constraint for the HDA problem.  
- Subsequent lines: records that look like
    H101_temperature:<val>, F101_temperature:<val>,
    F102_temperature:<val>, F102_deltaP:<val>,
    leads to Metric=<metric>.
    These entries exist only for parameter sets that passed validation.

2. The conversation stream
- If your previous proposal was invalid, the immediately-preceding message
    will come from ValidatorAgent and start with "Invalid, ...".
- Use the reason in that message (e.g. which limit was exceeded) when
    adjusting your next suggestion.

Objective (one of the following)
-------------------------------
- Higher yield, or
- Higher yield/cost, or
- Lower cost

Rules for every turn
---------------------
1. Parse the entire constraint_memory to understand long-term trends.
2. Look at the most recent VALID parameter set (last line in memory).  
Also check the last chat message:  
    - If it begins with "Invalid," treat your last increments as rejected.  
    - Shrink or reverse the offending increment so the result will fall
      inside its constraint window.  
    - Leave other increments unchanged unless you have evidence from
      history that a different direction is better.

3. Produce one Python dict literal (not JSON) called changes, e.g.

{'H101_temperature': -10,
 'F101_temperature': 5,
 'F102_temperature': 0,
 'F102_deltaP': -5000}

- All four keys must appear.  
- The increments are RELATIVE adjustments (can be negative, positive, or 0).  

4. If you judge no further improvement is possible, output exactly:
    TERMINATE
(uppercase, nothing else).
\end{lstlisting}

\section{Constraint Generation}

This section presents the outputs generated by the ContextAgent to demonstrate the framework's capability to produce realistic process constraints and technical understanding from minimal input descriptions. The results validate the agent's embedded domain knowledge and its ability to infer engineering-appropriate operating bounds.

\subsection{Process Understanding}

To evaluate the ContextAgent's process comprehension capabilities, we examine the technical overview it generated for the HDA process. The following output demonstrates the agent's ability to synthesize process knowledge, identify key operational considerations, and provide technically accurate descriptions of unit operations and their interactions.

\begin{lstlisting}[caption={ContextAgent’s Process Overview Generation Result}, label={lst:hda_process}]
### Hydro-De-Alkylation (HDA) Process - Block-by-Block Summary
1. Feed Preparation
   - Liquid toluene (Stream B) and make-up/recycle hydrogen (Stream A + recycle loop) enter Mixer M101 at 303 K and 350 kPa.
   - M101 provides homogeneous vapour-phase mixing without heat duty.
2. Charge Heater H101
   - The mixed feed is fired in Heater H101 to attain the high inlet temperature required for gas-phase hydrodealkylation.
   - The heater operates under essentially the same pressure as the upstream mixer (approx. 350 kPa) and raises the temperature into the catalytic reaction window (approx. 827-975 K) while remaining below the metallurgical limit of common furnace tubes (approx. 1000 K).
3. Reactor R101
   - An adiabatic, fixed-bed catalyst reactor effects the exothermic conversion:
     C6H5CH3 + H2 -> C6H6 + CH4.
   - Design basis sets single-pass toluene conversion at 75%.
   - Reactor effluent leaves only a few kelvin below the inlet (because the reaction is exothermic yet limited to 75% conversion) and at roughly 350 kPa.
4. High-Pressure Flash F101
   - The hot reactor effluent is partially condensed to 305-369 K at approx. 350 kPa with negligible pressure loss.
   - Vapour phase (rich in H2 and CH4) proceeds to the purge/recycle loop; liquid phase (benzene/toluene) remains for further separation.
5. Purge & Recycle Loop
   - Splitter S101 removes 20% of the flash vapour as purge to control inerts (chiefly CH4).
   - The remaining 80% is compressed isothermally in C101 back to the original system pressure of 350 kPa and returned to Mixer M101.
6. Low-Pressure Flash F102
   - The condensed liquid from F101 is throttled across a control valve, generating a pressure drop of 2 x 10^4 - 2 x 10^5 Pa (negative sign convention), well within the 250 kPa mechanical limit of the vessel.
   - At the lower flash pressure, benzene preferentially vaporises and exits overhead; the toluene-rich bottoms are recycled or sent to further recovery.
   - F102 temperature is maintained in the 306-366 K window to balance benzene recovery with heat-duty economy.
   - By specification, the attainable temperature in Heater H101 is completely independent of the pressure drop selected for F102.
7. Product Handling
   - Benzene overhead is condensed, stabilised, and routed to storage.
   - Toluene bottoms can be sent back to the reaction section or exported, depending on overall plant optimisation.
### Utilities & Controls (brief)
- Fired duty in H101 is trimmed to hold R101 inlet temperature.
- Level control in F101/F102 maintains phase disengagement.
- Pressure-override control on the purge stream prevents light-gas accumulation.
\end{lstlisting}

\subsection{Alternative Constraint Generation Approach Comparison}

To evaluate the importance of embedded domain knowledge in constraint generation, we conducted preliminary testing using a retrieval-augmented generation (RAG) approach that relied on external documents rather than the LLM's pre-trained knowledge. The RAG-based implementation encountered difficulties in generating complete constraint sets for all required variables, potentially due to document formatting inconsistencies or retrieval strategy limitations.  While these results suggest that embedded process knowledge may provide advantages for autonomous constraint generation, further investigation with curated industrial documents and refined RAG implementations would be needed to draw definitive conclusions about the merits of different knowledge integration approaches.

\section{Framework Convergence and Reproducibility Analysis}
To assess the convergence of the proposed optimization method, we conducted five independent runs for cost minimization using the same generated constraints and process overview.

\begin{figure}[H]
  \centering
  \includegraphics[width=0.9\textwidth]{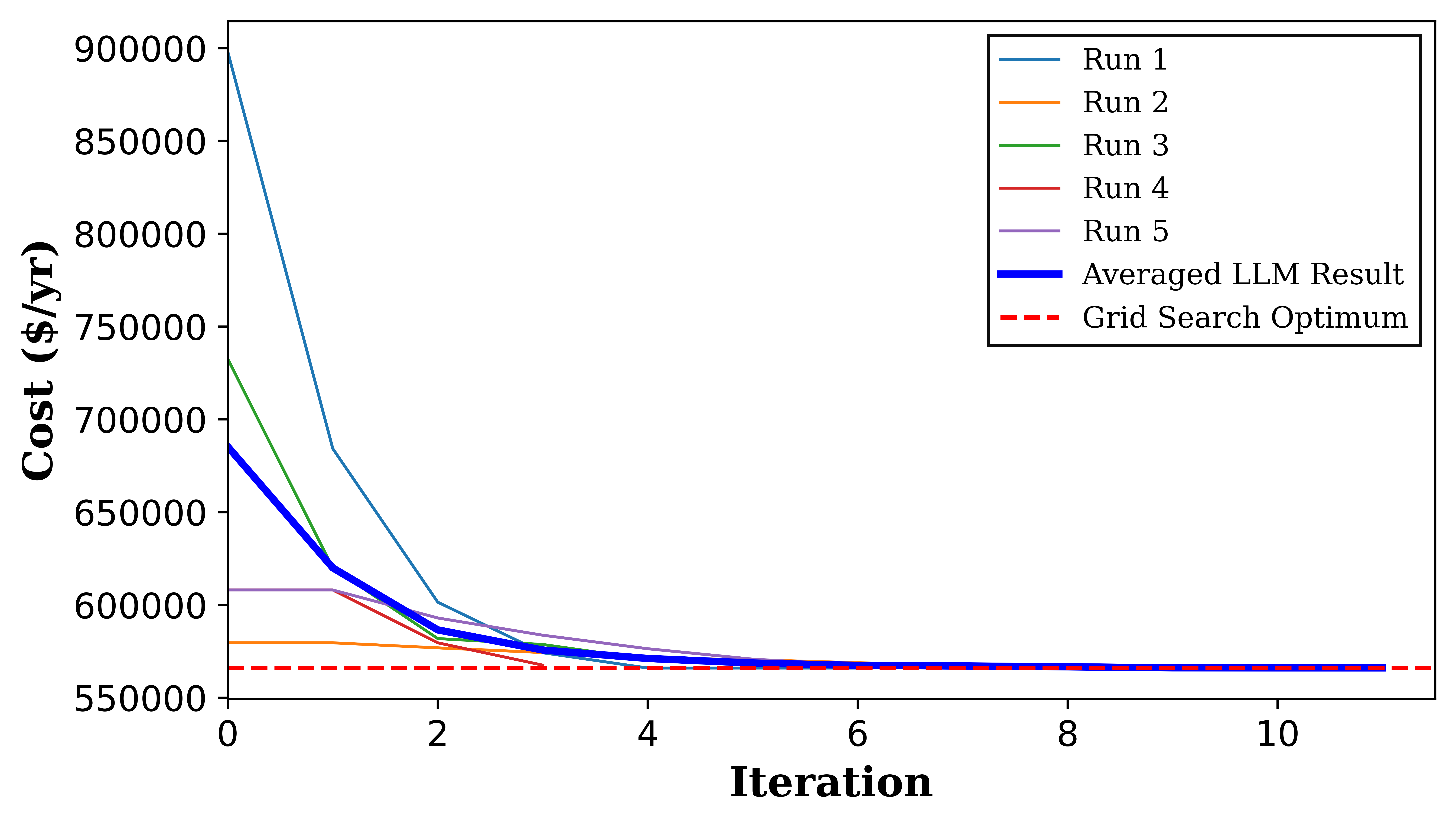}
\caption{Convergence of cost minimization across five independent runs using the October 2025 version of OpenAI's o3 model. Thin lines show individual runs; thick blue curve shows average trajectory across runs; red dashed line represents grid search's optimal solution of \$$5.660 \times 10^5$, serving as a benchmark for comparison with the LLM-based optimization approach.}
  \label{fig:convergence_cost}
\end{figure}

Figure~\ref{fig:convergence_cost} presents the cost evolution across these runs, comparing individual optimization trajectories against the average trajectory across all runs and the grid-search benchmark. Only validated, successfully simulated iterations are included in the figure, as the ValidationAgent filters infeasible proposals before they reach simulation. The results demonstrate rapid convergence, with the average trajectory decreasing sharply in early iterations before stabilizing near the benchmark solution. Notably, the divergence among cost trajectories decreases as optimization progresses, indicating that the framework achieves consistent and reproducible convergence behavior.

These convergence results were obtained using OpenAI's o3 model accessed in October 2025, which exhibits improved performance compared to results obtained in June 2025. The performance improvement may reflect model optimizations or updates between these testing periods.

\section{Baseline Method Overview}

IPOPT (Interior Point OPTimizer) is a gradient-based local optimizer designed to handle nonlinear programming problems (NLP). Mathematically, the NLP problem solved by IPOPT is expressed as\cite{wachter2006implementation}:
\begin{equation}
\eqalign{
    & \min_{x} \quad f(x) \cr
    & \text{s.t.} \quad g(x) \leq 0,\quad h(x) = 0,\quad x^L \leq x \leq x^U
}
\end{equation}
where \(f(x)\) is the objective function, \(g(x)\) and \(h(x)\) are inequality and equality constraints, and \(x^L, x^U\) represent variable bounds.  

For reliable convergence, IPOPT requires explicit constraints and at least first-order derivatives (exact or automatic). Providing twice-continuously-differentiable functions with accurate Hessian information improves robustness but is not strictly mandatory. Consequently, IPOPT can be less effective in early-stage problems where derivatives or complete constraints are unavailable. It is also vulnerable to getting stuck in local minima or convergence failures when faced with poorly defined spaces or discontinuous objectives.

\section{HDA Process Description and Optimization Setup}

\subsection{Process Specifications}

The hydrodealkylation (HDA) process converts toluene to benzene through catalytic reaction with hydrogen according to the reaction:

\begin{equation}
\text{C}_6\text{H}_5\text{CH}_3 + \text{H}_2 \rightarrow \text{C}_6\text{H}_6 + \text{CH}_4
\label{eq:hda_reaction}
\end{equation}

The process flowsheet (Figure in main text) consists of seven interconnected units: mixer (M101), heater (H101), reactor (R101), primary flash separator (F101), splitter (S101), compressor (C101), and secondary flash separator (F102). The process incorporates a recycle loop to maximize hydrogen utilization while managing methane accumulation through controlled purging.

The process operates with two fresh feed streams that establish the baseline material balance. Stream A provides the hydrogen feed containing 0.30 mol/s H$_2$ and 0.02 mol/s CH$_4$ at 303.2 K and 350 kPa, while Stream B supplies 0.30 mol/s toluene at identical temperature and pressure conditions. These feed specifications represent typical industrial feedstock compositions with the hydrogen-to-toluene ratio designed to meet stoichiometric requirements at the target conversion level.

Fixed process parameters define the fundamental design constraints that cannot be modified during operational optimization. The reactor operates at a fixed single-pass toluene conversion of 75\%, representing a design specification that balances reaction kinetics with equipment sizing requirements. The splitter maintains a constant split fraction of 20\% to purge and 80\% to recycle, establishing the recycle management strategy that prevents methane accumulation while maximizing hydrogen recovery. The compressor operates isothermally with a fixed outlet pressure of 350 kPa, restoring the vapor recycle stream to process operating pressure. The primary flash separator operates at approximately 350 kPa with negligible pressure drop across the unit.

The optimization framework manipulates four key operating parameters that directly influence process performance through their effects on reaction rates, separation efficiency, and energy consumption. The H101 outlet temperature serves as the reactor inlet temperature, controlling reaction kinetics and energy requirements for the exothermic heating duty. The F101 temperature determines the primary flash separation conditions, affecting the vapor-liquid split between light gases and heavy aromatics. The F102 temperature controls the secondary flash operating conditions for final product purification, influencing the benzene-toluene separation efficiency. The F102 pressure drop creates the driving force for enhanced separation by reducing the operating pressure and increasing the relative volatility between benzene and toluene.

\subsection{Objectives and Constraints}

Three optimization metrics are evaluated to capture different aspects of process performance. Specifically, the annual operating cost objective minimizes energy consumption through optimal utility management:

\begin{equation}
\text{Operating Cost} = \text{Heating Cost} + \text{Cooling Cost}
\label{eq:cost_objective}
\end{equation}

where heating costs account for the major energy consumers in the process:

\begin{equation}
\text{Heating Cost} = 2.2 \times 10^{-7} \cdot Q_{\text{H101}} + 1.9 \times 10^{-7} \cdot Q_{\text{F102}} + 63,931.475
\label{eq:heating_cost}
\end{equation}

and cooling costs include heat removal requirements:

\begin{equation}
\text{Cooling Cost} = 2.12 \times 10^{-8} \cdot |Q_{\text{F101}}| + 2.12 \times 10^{-8} \cdot |Q_{\text{R101}}|
\label{eq:cooling_cost}
\end{equation}
In Equation (4) and (5), $Q_{H101}$ (Heater H101), $Q_{F102}$ (Flash F102),$Q_{F101}$ (Flash F101) and $Q_{R101}$ (Reactor R101) represent the total annual heat duties (energy consumption) in units of Joules per year (J/yr). 
The heating cost coefficients reflect different utility requirements across process units. The H101 coefficient $(2.2 \times 10^{-7} \$/J)$ represents high-temperature steam costs for reactor preheating from ambient to reaction temperature, while the F102 coefficient $(1.9 \times 10^{-7} \$/J)$ reflects lower-grade heating utility requirements for flash separation optimization. The constant term $(63,931.475 \$/yr)$ accounts for fixed operating expenses including maintenance, labor, and baseline utility consumption. Cooling cost coefficients $(2.12 \times 10^{-8} \$/J)$ represent cooling water or refrigeration costs for heat removal from the primary flash separator (F101) and reactor temperature management (R101). The lower magnitude compared to heating costs reflects the typically lower cost of cooling utilities versus high-temperature heating media. The other two evaluation metrics are defined in the main manuscript. 

The optimization problem operates within physical and operational constraints that reflect thermodynamic feasibility and equipment limitations. Temperature limits are determined by vapor-liquid equilibrium requirements and equipment design ratings, typically ranging from 300-400 K for flash operations and 800-1000 K for reactor inlet conditions. The F102 pressure drop constraint range [-250,000, 0] Pa reflects equipment design limitations where excessive pressure drops can cause mechanical stress while positive values are thermodynamically impossible for expansion operations.

Flow rate constraints are bounded by the fixed feed specifications (0.30 $mol/s$ toluene, 0.30 $mol/s$ $H_2$, 0.02 $mol/s$ $CH_4$) and steady-state material balance requirements throughout the process network. Phase equilibrium constraints govern all vapor-liquid separations based on component volatilities, with benzene and toluene exhibiting significantly different vapor pressures that enable effective separation.

The reactor operates adiabatically $(Q = 0)$ with the fixed conversion constraint, eliminating reactor sizing as an optimization variable while allowing the exothermic heat of reaction $(\Delta H_{rxn} = -108 kJ/mol)$\cite{idaes_hda_flowsheet_2024} to contribute to maintaining reaction temperature. Despite the exothermic reaction, substantial external heating is required in H101 due to sensible heat requirements for elevating the feed mixture from 303.2 K to reaction temperature.

The recycle loop creates steady-state material balances that significantly influence overall process economics through their impact on raw material efficiency and separation requirements. Unreacted hydrogen is recovered and recycled through 80\% of the vapor stream from F101, while methane accumulation is controlled by directing 20\% of the vapor stream to purge. The fresh feed hydrogen flow rate of 0.30 mol/s exactly matches the stoichiometric requirements for complete conversion at the 75\% conversion level.

The material utilization efficiency depends critically on the balance between hydrogen recovery through recycling and methane removal through purging. This balance creates fundamental trade-offs between raw material costs and separation performance, as the purge stream removes valuable unreacted hydrogen along with the undesired methane byproduct. The steady-state operation requires that methane generation rates in the reactor equal methane removal rates in the purge stream, establishing equilibrium concentration levels throughout the recycle system.

\subsection{IDAES Implementation}

The process is modeled using the IDAES platform with ideal vapor-liquid equilibrium assumptions for all separation units. Component properties are based on standard correlations available in the IDAES property database, with the reactor kinetics represented through the fixed conversion constraint rather than detailed rate expressions. The simulation framework evaluates steady-state performance for each parameter set proposed by the LLM optimization agents, providing objective function values that guide the iterative optimization process. The IDAES implementation enables rapid evaluation of process performance while maintaining sufficient fidelity to capture the essential trade-offs between energy consumption, separation efficiency, and production capacity.

\section*{References}

\bibliographystyle{unsrt}
\bibliography{references}